\documentclass[conference, 10pt]{IEEEtran}

\usepackage[numbers,sort]{natbib}
\usepackage[dvipsnames]{xcolor}
\usepackage{graphicx}
\usepackage{amsmath} 
\usepackage{amssymb}
\usepackage{algorithm}
\usepackage[noend]{algpseudocode}
\usepackage{mathrsfs}
\usepackage{multirow,tabularx}
\usepackage{makecell}
\usepackage{dsfont}
\usepackage[T1]{fontenc}
\usepackage{booktabs}
\usepackage{adjustbox}
\usepackage{enumitem}
\usepackage{url}

\usepackage{tikz}
\usepackage{pgfplots}
\pgfplotsset{compat=1.14}
\definecolor{TolBlue}{HTML}{4477AA}
\definecolor{TolRed}{HTML}{CC6677}
\definecolor{TolYellow}{HTML}{DDCC77}

\newcommand{\Seq}{S}
\newcommand{\OSeq}{S'}
\newcommand{\RuleSet}{\mathcal{R}}
\newcommand{\Rule}{R}
\newcommand{\RAnt}{A}
\newcommand{\RCon}{C}
\newcommand{\RWgt}{w}
\newcommand{\concat}{\oplus}
\newcommand{\elem}{\sigma}
\newcommand{\Alphabet}{\Sigma}
\newcommand{\Model}{H}
\newcommand{\EmptyMod}{\RuleSet^{\emptyset}}
\newcommand{\ModFamily}{\mathcal{H}}
\newcommand{\lenS}{n}
\newcommand{\abs}[1]{\left\lvert#1\right\rvert}
\newcommand{\trule}[2]{#1 \Rightarrow #2}
\newcommand{\SingSet}{\Psi}
\newcommand{\OptimalWgt}{\hat{\vec{\RWgt}}}
\newcommand{\VectorWgt}{\vec{\RWgt}}
\newcommand{\CandidateSet}{\mathcal{C}}

\DeclareMathOperator{\pred}{predict}
\DeclareMathOperator{\activ}{active}
\DeclareMathOperator{\supp}{supp}
\DeclareMathOperator{\conf}{conf}
\DeclareMathOperator{\cgain}{gain}
\newcommand{\ip}{\mathit{ip}}
\newcommand{\frq}{f}

\newcommand{\trigg}[3]{#1 \vdash_#3 #2}
\newcommand{\appl}[3]{#1 \Vdash_#3 #2}
\newcommand{\predp}[3]{P_{#1,#2}(#3)}
\DeclareMathOperator{\cl}{L}
\newcommand{\algCossu}{\textsc{Cossu}}

\newcommand{\ignore}[1]{{}}

\newcolumntype{C}[1]{>{\centering\arraybackslash}p{#1}}
\renewcommand{\descriptionlabel}[1]{%
  \hspace{\labelsep}\normalfont\textit{#1}%
}
\newcommand{\spara}[1]{\smallskip\noindent \textbf{#1}}

\begin{document}
\newtheorem{exmp}{Example}

\title{Discovering Useful Compact Sets of Sequential Rules in a Long Sequence}

%

\author{
    \IEEEauthorblockN{
        Erwan Bourrand\IEEEauthorrefmark{1,2},
        Luis Gal\'{a}rraga\IEEEauthorrefmark{3},
        Esther Galbrun\IEEEauthorrefmark{4},
        Elisa Fromont\IEEEauthorrefmark{1},
        Alexandre Termier\IEEEauthorrefmark{1}}

    \IEEEauthorblockA{
        \IEEEauthorrefmark{1}
        \textit{Univ Rennes, IRISA UMR 6074, Rennes, France} \\
        \IEEEauthorrefmark{2}
        \textit{Advisor\_SLA, Inc.} \\
        \IEEEauthorrefmark{3}
        \textit{Inria RBA Rennes, France} \\
        \IEEEauthorrefmark{4}
        \textit{University of Eastern Finland} \\}

    \IEEEauthorblockA{
        \{luis.galarraga, elisa.fromont, alexandre.termier\}@irisa.fr}
}

\maketitle              

\begin{abstract} 
We are interested in understanding the underlying generation process for long sequences of symbolic events.  To do so, we propose \algCossu{}, an algorithm to mine small and meaningful sets of sequential rules. The rules are selected using an MDL-inspired criterion that favors compactness and relies on a novel rule-based encoding scheme for sequences. Our evaluation shows that \algCossu{} can successfully retrieve relevant sets of closed sequential rules from a long sequence.
Such rules constitute an interpretable model that exhibits competitive accuracy for the tasks of next-element prediction and classification.
\end{abstract}

\section{Introduction}
\label{sec:intro}

Long sequences of symbols are ubiquitous.
Examples include DNA sequences, server logs, traces of network packets, and even long texts. 
Discovering regularities in such sequences allows for a better understanding of the 
sequence generation process, and can be useful e.g., for diagnostic and prediction. 

Some models, such as LSTMs~\cite{Hochreiter97}, excel at detecting those regularities and use them for accurate prediction. On the downside, such models are hardly understandable by human users. Models based on pattern mining, in contrast, provide high-level human-readable descriptions of the underlying structure of the data. However, this interpretability typically comes at the cost of a restricted predictive power. 
We focus on the latter category of models, since we are interested in prediction but also in knowledge discovery. 

In the realm of pattern mining, {\em sequential rules}
are a well-established model for understanding and predicting sequential data.
Sequential rules take the form $\trule{A}{C}$, where both the antecedent $A$ and the consequent $C$ are sequences of items, e.g., events, nucleotides, or words.
A \emph{confidence} score is often attached to the rule to measure how likely it is to observe $C$ after $A$ in the sequence. For instance, the rule $\trule{A_1A_2A_3}{\textit{PowerFail}}$ with confidence $80\%$ tell us that there is a four in five chance of undergoing a power failure after having observed the sequence of alarms $A_1$, $A_2$ and $A_3$. This can be used both for prediction, i.e., to issue an early warning, and for diagnostic, i.e., to identify the link between the alarms and the power failure.

While sequential rule mining has been extensively studied in the literature \cite{fournier2017survey,association-rules-long-sequences,partially-ordered-sequential-rules,clofast,erminer}, two major issues remain. 
First, the vast majority of approaches consider a database of short sequences rather than a long sequence. While a long sequence can always
be split into a database, this incurs a loss of information at the boundaries.
Second, to the best of our knowledge, no approach for sequential rule mining addresses the so-called {\em pattern explosion}, the fact that millions of sequential rules may be produced due to the combinatorial search space. 

We propose the first sequential rule mining approach that takes as input a long sequence and outputs a compact set of rules.
The selection of the rules is based on the principle of Minimum Description Length (MDL), a paradigm that has proved effective for
model selection in pattern mining~\cite{krimp,sqs}. 
MDL approaches rely on an encoding scheme. Ours is presented in Section~\ref{sec:approach}, along with our algorithm to mine rules, in Section~\ref{sec:algo}. Experiments, presented in Section~\ref{sec:evaluation}, demonstrate the relevance of the rules found, as well as their predictive power for the tasks of next-element prediction, and classification for long sequences.

\section{Background}
\label{sec:background}
Our work stands at the crossroads of sequential and MDL-based pattern mining. 
Below we give brief overviews of these two fields, in turn.

\spara{Sequential Pattern Mining.}
Mining sequential patterns on data is a well-studied problem that has given rise to a plethora of methods 
accounting for different notions of patterns. 
The vast majority of approaches assume a database of sequences as input,
where sequences are ordered collections of itemsets. 
We 
refer the reader to the survey in~\cite{fournier2017survey} for a detailed account.  

Among the most recent works, Fumarola et al.~\cite{clofast} mine frequent closed sequences, i.e.\ sequences of maximal length such that any extension will inevitably have lower support. Closed sequences can alleviate pattern explosion to some extent as they are a subset of frequent sequences. Other approaches~\cite{rulegrowth,erminer,partially-ordered-sequential-rules} mine sequential rules of the form $\trule{\RAnt}{\RCon}$ where $\RAnt$ and $\RCon$ 
may be itemsets~\cite{rulegrowth,erminer} or sub-sequences~\cite{partially-ordered-sequential-rules} with the constraint that $\RCon$ occurs after $\RAnt$. Itemsets are generally easier to mine than sequences, however they are less expressive because they do not account for order in the appearance of items.  
Still, none of these methods is directly portable to long sequences as they all assume the data has been partitioned into small sequences (and hence often rely on a different definition of support). A few methods~\cite{association-rules-long-sequences,Gupta2006} can natively mine rules on long sequences. Unfortunately, such methods are not resilient to pattern explosion and are limited to association rules between itemsets, thus they do not capture the order between items within the antecedent and the consequent.

\spara{Mining patterns with an MDL criterion.}
\label{subsec:mdl}
The Minimum Description Length (MDL) principle~\cite{Grunwald:2007:MDL:1213810} 
is a criterion rooted in information theory that stipulates that the best model 
for a dataset is the model that compresses it best. 
For a dataset $D$ and a family of models $\ModFamily$, the best model for $D$, according to the (two-parts) MDL, is the one that minimizes the description length 
\begin{equation} \label{eq:mdl}
\Model^* = \arg\min_{\Model \in \ModFamily} \; \cl(\Model) + \cl(D | \Model)\,, 
\end{equation}
where $\cl$ denotes code length in bits. 
Equation~(\ref{eq:mdl}) strikes a balance between model complexity, as measured by $\cl(\Model)$,
and fitness to the data, as measured by $\cl(D | \Model)$. 
When applying MDL to pattern mining, a model is a collection of patterns, e.g.,\ itemsets~\cite{krimp}, sequences~\cite{sqs} 
or, in our case, sequential rules. One of the main ingredients of any MDL-inspired pattern mining approach is
the \emph{encoding scheme}, namely, the protocol to encode the data with the patterns 
and to encode the patterns themselves.
Once an encoding mechanism is in place, we can generate candidate sets of patterns, calculate the corresponding code lengths, 
and select the set resulting in the shortest code length.

\textsc{Krimp}~\cite{krimp} was one of the pioneering efforts to apply MDL to pattern mining, more specifically to itemset mining. 
Given a (large) collection of itemsets mined from the data, \textsc{Krimp} selects a small representative subset. The selected itemsets were also empirically shown to be effective for classification.
Recently, Fischer and Vreeken~\cite{Fischer19} proposed an approach to mine compact sets of rules from data without the sequential dimension. 
The approaches proposed in~\cite{BertensVS16,BhattacharyyaV17} extract sequences with gaps (but not rules) from univariate and multivariate long sequences. Unlike \textsc{Krimp}~\cite{krimp}, they combine the generation and selection of candidate patterns, which boosts efficiency and pattern quality.
Shokoohi-Yekta et al.~\cite{Shokoohi15} draw inspiration from the MDL principle to mine sequential rules from time-series discretized into an integer domain.
A mining approach tailored for the streaming setting is proposed in~\cite{swift}, but its compliance with the MDL principle is debatable.  


\section{Definitions and Notation}
A \emph{sequence} $\Seq$ of length $\lenS$ over an alphabet $\Alphabet$ is an ordered collection of $\lenS$ occurrences of symbols from $\Alphabet$.
We denote by $\Seq[i]$ the $i$-th element in $\Seq$ ($1 \leq i \leq \lenS$), and let $\Seq[i,j]$ denote 
the contiguous sequence $\langle \Seq[i] \dots \Seq[j] \rangle$ when $1 \leq i \leq j \le \lenS$ or the empty sequence $\emptyset$ otherwise.
For simplicity, we write the sequence $\langle \sigma_1, \sigma_2 \dots \sigma_k \rangle$ as $\sigma_1\sigma_2\dots \sigma_k$.
We denote the concatenation of two sequences by $\oplus$.

We say that sequence $\OSeq$ is a \emph{subsequence} of $\Seq$ (respectively $\Seq$ is a \emph{supersequence} of $\OSeq$), denoted as $S' \sqsubseteq S$, iff there exist $i$ and $j$ in $[1,\lenS]$ such that $\OSeq = \Seq[i,j]$. Then, we call $\Seq[i,j]$ a \emph{match} of $\OSeq$ in $\Seq$. More specifically, we denote by $\OSeq \sqsubseteq_i \Seq$ the fact that $\OSeq$ is a subsequence of $\Seq$ with a match starting at position $i$, i.e.\ $\OSeq=\Seq[i, i + \abs{\OSeq}-1]$, and by $\OSeq \sqsubseteq^j \Seq$ the fact that $\OSeq$ is a subsequence of $\Seq$ with a match ending at position $j$, i.e.\ $\OSeq=\Seq[j - \abs{\OSeq} + 1, j]$. 

The \emph{support} of a subsequence $\OSeq$ in $\Seq$, denoted by $\supp_{\Seq}(\OSeq)$, is the number of distinct matches of $\OSeq$ in $\Seq$, that is, $\supp_{\Seq}(\OSeq) =\abs{\{ i \in [1,\lenS]:\; \OSeq \sqsubseteq_i \Seq\}} =\abs{\{ j \in [1,\lenS]:\; \OSeq \sqsubseteq^j \Seq\}}$.
Finally, a sequence $\OSeq$ is \emph{closed} in $\Seq$ if there is no supersequence $\Seq'' \sqsupset \OSeq$ in $\Seq$ such that $\supp_{\Seq}(\Seq'') = \supp_{\Seq}(\OSeq)$.

\begin{exmp}{Sequences.}\label{ex:sequences}
\normalfont{
Given the sequences $\Seq = abceabcadeab$ and $\OSeq = abc$, $\OSeq \sqsubseteq_{1} \Seq$, $\OSeq \sqsubseteq_{5} \Seq$, $\OSeq \sqsubseteq^{3} \Seq$ and $\OSeq \sqsubseteq^{7} \Seq$ are all true, hence $\supp_{\Seq}(\OSeq) = 2$.
}
\end{exmp}

A \emph{rule} is a pair of sequences $(\RAnt, \RCon)$ over alphabet $\Alphabet$, such that $\RCon \neq \emptyset$. It is denoted as $\trule{\RAnt}{\RCon}$. $\RAnt$ and $\RCon$ are called the \emph{antecedent} and the \emph{consequent} of the rule, respectively.
A rule such that $\RAnt = \emptyset$ and $\abs{\RCon}=1$ is called a \emph{singleton rule}. The set of singleton rules over alphabet $\Alphabet$ is $\SingSet_{\Alphabet} = \{ \trule{\emptyset}{\langle \elem \rangle}:\;  \elem \in \Alphabet \}$. 

A rule $\Rule: \trule{\RAnt}{\RCon}$ \emph{triggers} on sequence $\Seq$ at position $i$, 
denoted as $\trigg{\Rule}{\Seq}{i}$, iff the antecedent has a match in $\Seq$ ending at position $i$, i.e., iff $\RAnt \sqsubseteq^i \Seq$. 
The rule $\Rule$ \emph{applies} on sequence $\Seq$ at position $i$, denoted by $\appl{\Rule}{\Seq}{i}$, iff the antecedent has a match in $\Seq$ ending at position $i$ and the consequent has a match in $\Seq$ starting at position $i+1$, i.e.\ if $\RAnt \sqsubseteq^i \Seq$ and $\RCon \sqsubseteq_{i+1} \Seq$.
Clearly, $\appl{\Rule}{\Seq}{i}$ implies $\trigg{\Rule}{\Seq}{i}$.

The \emph{support} of a rule is the number of times it applies in the sequence, whereas its \emph{confidence} is the ratio of the number of times the rule applies to the number of times the rule triggers:
\[\supp_{\Seq}(\Rule) = \abs{\{i \in [1,\lenS]:\; \appl{\Rule}{\Seq}{i} \}} \]

and
\[\conf_{\Seq}(\Rule) = \frac{\abs{\{i \in [1,\lenS]:\; \appl{\Rule}{\Seq}{i}\}}}{\abs{\{i \in [1,\lenS]:\; \trigg{\Rule}{\Seq}{i}\}}}\,.\]

\begin{exmp}{Rules.}\label{ex:rules}
\normalfont{
In sequence $\Seq$ from Example~\ref{ex:sequences}, rule $\Rule : \trule{ab}{c}$ has support $\supp_{\Seq}(\Rule) = 2$ and confidence $\conf_{\Seq}(\Rule) = 2/3$, because $R$ triggers at positions $1$, $5$ and $11$ but only applies at positions $1$ and $5$. 
}
\end{exmp}

Given a sequence $\Seq$ up to position $m$ and a rule $\Rule: \trule{\RAnt}{\RCon}$, we say that $\Rule$ is \emph{active} at stage $j$ ($0 \le j < \abs{\RCon}$) and \emph{predicts} $\RCon[j+1]$, if $\RAnt\concat\RCon[1,j] = \Seq[m - \abs{\RAnt} -j, m]$. This corresponds to cases where the antecedent of $\Rule$ is followed by the first $j$ elements of its consequent (empty in case $j=0$) in $\Seq$.
Formally, we define the predicates $\activ(\Rule, \Seq, j)$, to indicate that rule $\Rule$ is active on sequence $\Seq$ at stage $j$, and $\pred(\Rule, \Seq, j)$, to return $\RCon[j+1]$ if $\activ(\Rule, \Seq, j)$ is true. 


\section{Encoding Scheme}
\label{sec:approach}
To illustrate our encoding scheme, we resort to a sender-receiver metaphor, where the sender first transmits a set of rules and their corresponding weights, 
followed by a sequence of code words, one for each element of the sequence, in such a way that the receiver can reconstruct the original sequence.

\spara{Encoding a Sequence via Sequential Rules.}
Assume we want to predict the next element after $cabba$, based on two rules 
$\Rule_1: \trule{abba}{b}$ and $\Rule_2: \trule{ba}{cd}$. 
If the confidence of $\Rule_1$ is higher than the confidence of $\Rule_2$, 
we would guess that $b$ is more likely than $c$ to appear next.
We rely on this simple principle to encode a sequence element by element using a set of sequential rules.
Assuming the receiver knows the set of rules and their confidence, 
as represented by weights, as well as the history of the sequence decoded so far, they can compute a probability distribution 
over the possible next elements. 
Hence, the sender can represent the next element as a code word chosen according to this probability distribution. 
If the set of rules effectively produces a probability distribution concentrated 
on the element that indeed occurs next, the code word for that element will be short, allowing 
for a concise representation of the input sequence.

Given a sequence $\Seq$ of $m$ elements seen so far, and a set of weighted rules $(\trule{\RAnt_{\Rule}}{ \RCon_{\Rule}}, \RWgt_{\Rule})$, such that 
$\RAnt_{\Rule}$, $\RCon_{\Rule}$, and $\RWgt_{\Rule}$ are respectively the antecedent, the consequent, and the weight of rule $\Rule$, 
our goal is to compute the probability of the next element given this set of rules  -- denoted by $\RuleSet$. This amounts to computing a probability distribution over $\Alphabet$.
For each possible next element $\elem \in \Alphabet$, 
we sum the weights of the active rules that predict $\sigma$ and divide by the sum of weights of all active rules:
\[\predp{\RuleSet}{\Seq}{\elem} = \frac{ \sum_{(\Rule,j) \in \mathcal{A}_{\Seq,\elem}} \RWgt_{\Rule}}{\sum_{(\Rule,j) \in \mathcal{A}_{\Seq}} \RWgt_{\Rule}}\] 
where
$\mathcal{A}_{\Seq} = \{ (\Rule, j),  (\Rule, \RWgt_{\Rule}) \in \RuleSet \text{ s.t.}\,\activ(\Rule, \Seq, j)\}$,
and 
$\mathcal{A}_{\Seq,\elem} = \{ (\Rule, j) \in \mathcal{A}_{\Seq} \text{ s.t.}\, \pred(\Rule, \Seq, j) = \elem\}$.
A full sequence $\Seq$ of length $n$ is transmitted element by element. At each step $m$, we encode the next symbol $\Seq[m]$ with a code word chosen according to the probability assigned to it based on $\RuleSet$ and the portion of the sequence seen so far, $\Seq[1,m-1]$.
Hence, the overall code length for the full sequence is 
\begin{equation} \label{eq:sequence-length}
\cl(\Seq | \RuleSet) = \sum_{m \in [1,n]} -\log_2\big( \predp{\RuleSet}{\Seq[1,m-1]}{\Seq[m]}\big).
\end{equation}

The receiver, having the same collection of rules $\RuleSet$ and previous elements of the sequence, can perfom the same computation to dynamically reconstruct the code 
at his end, and decode the transmitted element, using an agreed canonical order on the alphabet to break ties if necessary.

To ensure comparability, the models must achieve \emph{lossless} compression. 
For this reason, by construction our models always contain the singleton rule set $\SingSet_\Alphabet$. 
This way, every symbol of the alphabet receives a non-zero probability of occurrence at any step, and can hence be transmitted.
In fact, the singleton rule set constitutes the basis of the simplest possible model, what we call the \emph{empty model} and denote by $\EmptyMod$, 
since it does not contain any proper rule. 
To each rule $\Rule_{\elem} \in \SingSet_\Alphabet$ we associate a weight equal to the \emph{background probability} 
of the predicted element, estimated as its frequency in the sequence, $f_{\elem} = \abs{\{ i \in [1,\lenS]: S[i] = \elem \}}/\lenS$.


\spara{Encoding the Rules.}
\label{subsubsec:encoding-the-rules}
For the receiver to have access to the rules $\RuleSet$, the sender must transmit the antecedents, consequents, and weights at the start of the exchange.
The antecedent and consequent are subsequences, hence they can be encoded simply by stating the length of the sequence then listing their elements using codes of length $-\log_2(f_{\elem})$, in order.
The weight, on the other hand, is a real number of bounded precision in $(0, 1)$. Each weight is written as a finite list of decimals that we encode by applying universal coding to the integer value obtained by listing these decimals in reverse order. Put differently, we encode weight $w=0.d_1d_2d_3...d_k$ with a code word of length $\cl_{D}(w) = \cl_{\mathbb{N}}(d_k...d_3d_2d_1)$.
$\cl_{\mathbb{N}}(z)$ is the length of the code word assigned by the universal code to integer $z$, which is such that $\cl_{\mathbb{N}}(z) = \log_2^*(z) + \log_2(c_0)$, with $c_0$ a constant adjusted to ensure the Kraft inequality is satisfied.
This penalizes weights by the number of significant digits they contain rather than by their value.
Overall, the code length for a rule $\Rule$ is the sum of the lengths of the code words for the antecedent, the consequent and the weight, respectively:

\begin{small}
\[
\cl(\Rule) = \left[ \cl_{\mathbb{N}}(\abs{\RAnt_{\Rule}}+1)  + \sum_{\sigma \in \RAnt_{\Rule}}-\log_2(\frq_\sigma) \right] \] \[\qquad 
+ \left[ \cl_{\mathbb{N}}(\abs{\RCon_{\Rule}}) + \sum_{\sigma \in \RCon_{\Rule}}-\log_2(\frq_\sigma) \right] + \cl_{D}(\RWgt_{\Rule})\,.
\]
\end{small}

\noindent We encode a rule table by stating the number of rules and then listing them
\begin{equation}\label{eq:table-length}
\cl(\RuleSet) =  \cl_{\mathbb{N}}(\abs{\RuleSet}) + \sum_{\Rule \in \RuleSet} \cl(\Rule)\,.
\end{equation}

\section{The \algCossu{} algorithm}
\label{sec:algo}
We now introduce an algorithm to mine a set of sequential rules that compresses the input sequence well under the encoding scheme presented in Section~\ref{sec:approach}.
Finding such a set of rules is intractable in practice, thus we resort to heuristics.

Our \algCossu\,(COmpact Sets of Sequential rUles) algorithm is outlined in Algorithm~\ref{alg:cossu}.
Given a sequence $\Seq$ over alphabet $\Alphabet$ as input, \algCossu{} returns a set of rules $\RuleSet$ in two phases.
First, \emph{rule construction} (lines~\ref{alg:generate_start}--\ref{alg:generate_end}) generates a collection of candidate rules $\CandidateSet$, which are then evaluated during \emph{rule selection} (lines~\ref{alg:select_start}--\ref{alg:select_end}). 

\spara{Rule Construction.} \algCossu{} starts by extracting closed frequent subsequences by applying an off-the-shelf sequence miner \cite{Tatti:2012:MCS:2205934.2205997} to the input sequence $S$ with a minimum support threshold of $2$.
Sequential rules are then generated from the sequences by considering all possible partitionings into an antecedent and a non-empty consequent.
We use the \emph{compression gain} as a rough estimate of the individual ability of a rule to compress the sequence:

\begin{small}
\begin{equation}
    \cgain(\Rule) = \conf_{\Seq}(\Rule) \cdot \supp_{\Seq}(\Rule) \cdot \cl(\RCon_{\Rule}) - (\cl(\RAnt_{\Rule}) + \cl(\RCon_{\Rule}))\;.
  \end{equation}
\end{small}
This score puts in balance the potential benefit and cost of adding the rule to the rule set. The benefit depends on how often the consequent is correctly predicted (first term), whereas the cost is the code length of the rule's antecedent and consequent (second term). The latter term constitutes a lower bound on the cost, with a weight yet undetermined.
At the end of this first phase, the set of candidates consists of those rules that have a strictly positive compression gain.

\spara{Rule Selection.}
\algCossu{} resorts to a greedy strategy to build the rule set $\RuleSet$. Initially, $\RuleSet$ consists of the singleton rules (line~\ref{alg:select_start}). 
The algorithm then processes the candidate rules in $\CandidateSet$ by order of decreasing compression gain, i.e.\ from most promising to least promising (line~\ref{alg:select_loop}).
The candidate rule is tentatively added to the rule set and the weights are re-adjusted (line~\ref{alg:ruleset_adjust}). If the resulting rule set yields better compression than the current one, the current set is replaced (line~\ref{alg:ruleset_replace}) and the algorithm goes into a pruning loop. Otherwise, the rule is discarded and the algorithm moves on to the next candidate. The pruning loop (lines~\ref{alg:prune_start}--\ref{alg:prune_end}) checks whether the newly added rule makes any of the previously incorporated non-singleton rules obsolete. To do so, it  tentatively removes each of the rules in turn and checks whether the compression improves as a result, in which case the rule is permanently removed from $\RuleSet$.

\begin{algorithm}[tb]
\begin{small}
\caption{\algCossu{}: finding a compact set of sequential rules}
\label{alg:cossu}
\begin{algorithmic}[1]
\Require{A long sequence $\Seq$ of elements over an alphabet $\Alphabet$} 
\Ensure{A compact set of sequential rules $\RuleSet$}
    \State $\mathcal{S} \gets \text{MineClosedSequences}(S)$
    \label{alg:generate_start}
    \State $\CandidateSet \gets \{ \Rule \in \text{PrepareRules}(\mathcal{S}), \Rule \not\in \SingSet_{\Alphabet} \text{ and } \cgain(\Rule) > 0 \}$\label{alg:generate_end}
    \State $\RuleSet \gets \EmptyMod$, with adjusted weights $\OptimalWgt = \arg\min_{\vec{\RWgt}} \cl(\Seq|\EmptyMod)$ \label{alg:select_start} 
    \For{$\Rule \in \CandidateSet\text{, ordered by decreasing }\cgain(\Rule)$ \label{alg:select_loop}}
    \State $\RuleSet' \gets \RuleSet \cup \{ \Rule \}$, \newline \hspace*{1.5cm} with re-adjusted weights $\OptimalWgt = \arg\min_{\vec{\RWgt}} \cl(\Seq|\RuleSet')$ \label{alg:ruleset_adjust}    
    \If{$\cl(\RuleSet', \Seq) < \cl(\RuleSet, \Seq)$}
	\State $\RuleSet \gets \RuleSet'$ \label{alg:ruleset_replace}  
	\For{$\Rule' \in \RuleSet \setminus \EmptyMod$\label{alg:prune_start}}
	  \If{$\cl(\RuleSet \setminus \{\Rule'\}, \Seq) \leq \cl(\RuleSet, \Seq)$}
            \State $\RuleSet \gets \RuleSet \setminus \{ \Rule' \}$ \label{alg:prune_end}\label{alg:select_end}
	  \EndIf
     \EndFor
   \EndIf
   \EndFor
   \State \Return $\RuleSet$
\end{algorithmic}
\end{small}  
\end{algorithm}

\spara{Adjusting Rule Weights.}
Given a set of selected rules, \algCossu{} must determine a suitable vector of weights $\vec{\RWgt}$ associated to the rules, 
so as to optimize the code length of both the model $\cl(\RuleSet)$ (complexity) and the sequence $\cl(\Seq|\RuleSet)$  (fit).   
Because optimizing both aspects concurrently is very challenging, we assume that the code length of the weights is fixed, by fixing the floating point precision of the weights, and focus on the problem of minimizing $\cl(\Seq|\RuleSet)$, i.e.\ solving
\[\arg\min_{\VectorWgt} \sum_{m \in [1,n]} -\log_2\big( \predp{\RuleSet}{\Seq[1,m-1]}{\Seq[m]}\big) \;.\]
We do so greedily, by adjusting the weight of each rule in turn, while keeping the value of the other weights fixed.
The minimization resorts to the golden section search algorithm~\cite{golden-section-search}, initializing the weights to $1$.
In the end, we scale all weights to ensure that their values lie in the interval $(0, 1)$.
Designing an algorithm that is able to optimize both $\cl(\RuleSet)$ and $\cl(S|\RuleSet)$ is a major direction for future work.



\section{Evaluation}
\label{sec:evaluation}
We evaluate our proposed algorithm on both synthetic and real-world datasets. 
In particular, we evaluate the quality of the rules learned by \algCossu{} on real-world data, by using them as features in two classical machine learning tasks on long sequences: next event prediction and classification. While \algCossu{} is not designed specifically for these tasks, we show it still yields acceptable performance and reports rules that capture useful regularities in the data. 


\subsection{Evaluation on synthetic data}
\label{ssec:exp_synth}
To study the behaviour of our algorithm in controlled settings, we generate synthetic random sequences of symbols and inject patterns that match a set of hand-crafted sequential rules. Then, we verify whether \algCossu{} is able to recover the planted rules.

\spara{Data generation and experimental setting.}
%
The data generation process constructs an initial sequence $S$ by drawing $n$ symbols from a distribution $\mathscr{D}$ on an alphabet $\Sigma$. We then insert a set $\mathscr{T}$ of non-trivial rules (target) in the data. For each rule $R \in \mathscr{T}$, 
and with a probability equal to $\ip$, we insert the consequent of the rule after every match of the antecedent in $S$. This makes the sequence $S$ longer, thus we truncate it to keep only the $n$ first elements.  

By default $n=5000$, $\Sigma = \{A, B, C, D, E\}$, $\mathscr{R} = \{ A \rightarrow B\}$, with insertion probability $\ip=50\%$, and $\mathscr{D}$ is an equiprobable distribution on the elements of $\Sigma$. In each experiment we vary one of these parameters and leave the others at their default value. 

We highlight that because of the data and rule generation strategies, 
the insertion probability of a rule is not equivalent to its actual confidence $c$ 
in the final sequence. Their relationship is given by the expression $c = \ip + (1- \ip) \cdot P(B|A)$, where $P(B|A)$ is
the confidence of $A \rightarrow B$ in the original sequence. 

Our evaluation metric is the \emph{hit rate}, defined as the percentage of times \algCossu{} retrieves the exact set of rules that were introduced, which we call the gold standard. We report the hit rate across $100$ different sequences. 

\begin{table}[tb]
\centering
\caption{Hit rate and average number of candidate rules $\abs{\CandidateSet}$ for different insertion probabilities ($\ip$) of the single rule $A\rightarrow B$.
}
\begin{tabular}{@{\hspace{1pt}}l@{\hspace{.5em}}r@{\hspace{5pt}}r@{\hspace{5pt}}r@{\hspace{5pt}}r@{\hspace{5pt}}r@{\hspace{5pt}}r@{\hspace{5pt}}r@{\hspace{5pt}}r@{\hspace{5pt}}r@{\hspace{5pt}}r@{\hspace{1pt}}r@{\hspace{1pt}}}
\toprule
$\ip$ 	& 5\%  & 10\%  & 20\%  & 30\%  & 40\%  & 50\% & 60\%  & 70\%  & 80\%   & 90\%  & 100\% \\ \midrule
Hit rate (\%)  & 3	 & 75  & 100   & 100  & 100  & 100 & 100 & 100 & 100 & 94  & 100 \\ 
$\abs{\CandidateSet}$ ($10^3$)	& 13.7 & 13.7 & 13.8 & 14.0  & 14.2 & 14.5 & 14.8 &  15.1  & 15.5 & 15.8 & 16.2 \\ \bottomrule 
\end{tabular}
\label{tab:synthir}
\end{table}

\spara{Different rule insertion probability.}
\label{insertrate}
\ignore{The \emph{insertion probability} of a rule $A\rightarrow B$ is the probability to insert consequent $B$ in a generated sequence (after $A$) knowing that we already found $A$ in the sequence. 
Note that, because of the data and rule generation strategies, the insertion probability of a rule $A\rightarrow B$ is not equivalent to the confidence $c = \frac{P(A \,\cap\, B)}{P(A)}$ of the rule in the final sequence.
Indeed, the confidence of the rule comes from three elements: 1) the $B$ that we insert after $A$s found in the sequence that were not followed by $B$s 2) the $A B$ subsequences already in the sequence before insertion and where we do not add a supplementary $B$, 3) to which we must subtract the pre-existing $A B$ were the insertion added a supplementary $B$ anyway.
It follows that the final confidence of an inserted rule $A\rightarrow B$ is actually:
$$c = Insertion Probability + (1- Insertion Probability) * P(B)$$ 
}
In Table \ref{tab:synthir} we show the hit rate and average number of rules found by \algCossu{}
for different insertion probabilities when inserting a single rule.
Except when $\ip < 20\%$, \algCossu{} always achieves a hit rate of between $90\%$ and $100\%$. For low insertion rates, the confidence of the target rule is closer to the background probabilities of the 5 symbols. This makes \algCossu{} unable to distinguish the rules from the background and explains the low hit rates ($\leq 75\%$). 


\begin{table}[tb]
\centering
\caption{Hit rate and average number of candidate rules $\abs{\CandidateSet}$ for different alphabet sizes ($\abs{\Alphabet}$), no rule/one rule.}
\begin{tabular}{@{\hspace{1pt}}l@{\hspace{5pt}}r@{\,/\,}l@{\hspace{5pt}}r@{\,/\,}l@{\hspace{5pt}}r@{\,/\,}l@{\hspace{5pt}}r@{\,/\,}l@{\hspace{5pt}}r@{\,/\,}l@{\hspace{5pt}}r@{\,/\,}l@{\hspace{1pt}}}
\toprule
$\abs{\Alphabet}$ 	& \multicolumn{2}{c}{2}  & \multicolumn{2}{c}{3}  & \multicolumn{2}{c}{5}  & \multicolumn{2}{c}{10}  & \multicolumn{2}{c}{50}  & \multicolumn{2}{c}{100}   \\ \midrule
 Hit rate (\%)  & 100 & 42 & 100 & 100 & 100 & 100  & 100 & 100  & 100 & 100  & 100 & 100 \\ 
 $\abs{\CandidateSet}$ ($10^3$) & 54.2 & 58.3 & 26.1 & 28.0 & 13.6 & 14.5 & 7.0 & 7.3 & 3.3 & 3.3 & 1.9 & 1.9       \\ \bottomrule
\end{tabular}
\label{tab:synthalpha}
\end{table}

\spara{Different alphabet size.}
In Table \ref{tab:synthalpha}, we change the size of the alphabet $\Alphabet$ when generating the original sequence.
We can see that when no rule is inserted, \algCossu{} does not return any rule either even if the number of possible candidate rules $\abs{\CandidateSet}$ is often huge (and would have been output without the MDL constraints). The hit rate is only $42\%$ when there are only 2 symbols in the alphabet and we insert one single rule ($A\rightarrow B$). However, when $\abs{\Sigma}> 2$, \algCossu{} is able to find the inserted rule (and only this rule). 

\begin{table}[tb]
\centering
\caption{Hit rate and average number of candidate rules $\abs{\CandidateSet}$ for different sequence lengths, no rules/one rule.}
\begin{tabular}{@{\hspace{1pt}}l@{\hspace{5pt}}r@{\,/\,}l@{\hspace{5pt}}r@{\,/\,}l@{\hspace{5pt}}r@{\,/\,}l@{\hspace{5pt}}r@{\,/\,}l@{\hspace{5pt}}r@{\,/\,}l@{\hspace{5pt}}r@{\,/\,}l@{\hspace{1pt}}}
\toprule
 $\abs{S}$ & \multicolumn{2}{c}{100}  & \multicolumn{2}{c}{500}  & \multicolumn{2}{c}{1000}  & \multicolumn{2}{c}{2000}  & \multicolumn{2}{c}{5000}  & \multicolumn{2}{c}{10000}  \\ \midrule
Hit rate (\%)  & 100 & 16  & 100 & 99  & 100 & 100  & 100 & 100  & 100 & 100  & 100 & 100 \\ 
$\abs{\CandidateSet}$ ($10^3$) & 0.1 & 0.1 & 1.0 & 1.0 & 2.2 & 2.3 & 5.8 & 5.1 & 13.7 & 14.5 & 29.8 & 31.5 \\ \bottomrule
\end{tabular}
\label{tab:synthsequlength}
\end{table}

\spara{Different sequence length.}
In Table \ref{tab:synthsequlength} we show the hit rate of \algCossu{} when the sequence length varies. We can observe that if the sequence is too short ($\abs{S} \leq 500$) the rule $A\rightarrow B$ inserted is not occurring enough to be retrievable by our MDL-based algorithm: On average, for $\abs{S}=100$ the antecedent $A$ appears $20$ times so the subsequence $A B$ is present around 10 times in the sequence and hence does not constitute a useful regularity.
When the sequence becomes long enough ($\abs{S} > 500$), \algCossu{} correctly finds the inserted rule only.

\begin{table}[tb]
\caption{Hit rate and average number of candidate rules $\abs{\CandidateSet}$ for different inserted rule size (each time, only one rule is inserted in the generated sequence, with $\ip=0.5$).
\label{tab:synthrule1}}
\begin{tabular}{@{\hspace{1pt}}l@{\hspace{.2cm}}r@{\hspace{10pt}}r@{\hspace{10pt}}r@{\hspace{10pt}}r@{\hspace{1pt}}} 
\toprule
Injected rule & $A \rightarrow B$ & $A \rightarrow BC$ & $A \rightarrow BCD$ & $AB \rightarrow C$ \\
\midrule
Hit rate (\%)  & 100 & 100 & 100 & 100  \\ 
$\abs{\CandidateSet}$ ($10^3$) & 14.475 & 15.510 & 16.500 & 13.798
\\ [1.em]
Injected rule & $AB \rightarrow CD$ & $AB \rightarrow CDE$ & $ABC \rightarrow D$ & $ABC \rightarrow DE$ \\
\midrule
Hit rate (\%)   & 100 & 100 & 43 & 87 \\ 
$\abs{\CandidateSet}$ ($10^3$) & 13.956 & 14.155 & 13.655 & 13.703 \\
\bottomrule
\end{tabular}
\end{table}

\spara{Different rule size.}
In Table \ref{tab:synthrule1}, we study the behaviour of \algCossu{} when there are more than one symbol in the antecedent and/or consequent parts of the inserted rule. Note that the longer the antecedent is, the smaller the chances for the rule to occur in the sequence are, and the less likely \algCossu{} will need the rule to compress the sequence. On the contrary, the longer the consequent is (which is inserted each time the antecedent is present in the original data), the more likely \algCossu{} will select the rule to compress the data (esp.\ for short antecedents). The hit rate is $100\%$ for all rules where the size of the antecedent is lower than $3$ symbols. However, it is lower ($43\%$) when the antecedent size is greater or equal to $3$ symbols but again better ($87\%$) when the consequent of the rule is longer.


\spara{Different target rule set size.}
In Table \ref{tab:synthrule2} we study the behaviour of \algCossu{} when more than one rule is inserted in the synthetic sequence. Note that $A \rightarrow B \rightarrow C$ means that rules $A \rightarrow B$ and $B\rightarrow C$ were inserted. Recall also that the lower the number of symbols $\abs{\Sigma}$, the more difficult it is for \algCossu{} to find the rule in the data. We can see that \algCossu{} is stable, as it is able to retrieve any number of inserted rules in any configuration. In the difficult case where $\abs{\Sigma} =3$ (with 2 rules), \algCossu{} adds an additional rule $\emptyset \rightarrow BC$ in 4 of the runs as a way to assign a more precise weight to the rules. 


\begin{table}[tb]
\caption{Hit rate and average number of candidate rules $\abs{\CandidateSet}$ for different alphabet size and different inserted rules (with $\ip=0.5$ for all the inserted rules).\label{tab:synthrule2}}
\centering
\begin{tabular}{@{\hspace{1pt}}l@{\hspace{10pt}}r@{\hspace{10pt}}r@{\hspace{10pt}}r@{\hspace{1pt}}}
\toprule
Injected rules  & Hit rate (\%) & $\abs{\Alphabet}$ &  $\abs{\CandidateSet}$ ($10^3$) \\
\midrule
$A \rightarrow B \,;\, C \rightarrow D$  			&  100 & 4 & 19.865 \\
$A \rightarrow B \,;\, C \rightarrow D \,;\, E \rightarrow F$   		&  100 & 6 & 12.957 \\
$A \rightarrow B \rightarrow C$ 	&  96 & 3 & 30.405 \\
$A \rightarrow B \rightarrow C \rightarrow D$ 	&  100 & 4 & 21.880 \\
$A \rightarrow B \rightarrow C \rightarrow D \rightarrow E$ 	&  100  & 5 & 17.647 \\[.2em]
$ A$ \rotatebox{-15}{$\rightarrow$}  		 &  $100$ &  $7$ & $12.977$ \\
$ B \rightarrow E$ \rotatebox{-25}{$\rightarrow$}   &   &   & \\
$ D \rightarrow F \rightarrow G$   		 &  &   & \\
$ C$ \rotatebox{15}{$\rightarrow$}      &  &  & \\
\bottomrule
\end{tabular} 
\end{table}

\spara{Different alphabet distribution.}
Finally,  in Table \ref{tab:synthdistri} we evaluate the behaviour of \algCossu{} when the discrete probability distribution of the symbols in the generated data (before the rule insertion) is not uniform. 
We consider distributions where a symbol (the antecedent $A$ or the consequent $B$ of the inserted rule $A \rightarrow B$) is significantly more (denoted ``strong'') or less (denoted ``weak'') represented compared to the other symbols. In the ``increasing'' case, the probabilities of the two symbols A and B concerned by the inserted rule are lower than the probabilities of the other symbols: this makes it easier for \algCossu{} to retrieve the rule (hence the $99\%$ hit rate). In the ``decreasing'' case, the base probability of B is already high and the additional occurrences produced by the rule insertions in the data are often negligible compared to the original number of occurrences of B (hence the $72\%$ hit rate).
We can observe that, for the most unbalanced distribution named ``strong antecedent'' (resp. ``strong consequent'') in the table, the hit rate drops to $0\%$ (resp. $42\%$). In the first case the rule systematically retrieved is $\emptyset \rightarrow AB$ instead of $A \rightarrow B$. In the second case the base probability of the symbol $B$ is already high and, as discussed before, the rule cannot be easily spotted. Overall, this shows that \algCossu{} is sensitive to distributions with a particularly high entropy. This is further explored on real-world data in the following sections. 

\begin{table}[tb]
\caption{Hit rate and average number of candidate rules $\abs{\CandidateSet}$ for different alphabet distributions when inserting rule $A \rightarrow B$. \label{tab:synthdistri}}
\centering
\begin{tabular}{@{\hspace{1pt}}l@{\hspace{5pt}}r@{\hspace{5pt}}r@{\hspace{5pt}}r@{\hspace{5pt}}r@{\hspace{5pt}}r@{\hspace{3pt}}r@{\hspace{2pt}}r@{\hspace{1pt}}}
\toprule
Distribution & P(A) & P(B) & P(C) & P(D) & P(E) & Hit rate (\%) & $\abs{\CandidateSet}$ ($10^3$) \\ 
\midrule
Uniform & $1/\phantom{1}5$  & $1/\phantom{1}5$ & $1/\phantom{1}5$ & $1/\phantom{1}5$ & $1/\phantom{1}5$ &  $100$ & 14.5 \\
Increasing & $1/15$  & $2/15$ & $3/15$ & $4/15$ & $5/15$ &  $99$ & 15.8 \\
Decreasing & $5/15$  & $4/15$ & $3/15$ & $2/15$ & $1/15$ &  $72$ & 17.3 \\
Weak ante.\ & $1/41$  & $10/41$ & $10/41$ & $10/41$ & $10/41$ &  $98$ & 16.2 \\
Weak cons.\ & $10/41$  & $1/41$ & $10/41$ & $10/41$ & $10/41$ &  $84$ & 16.0 \\
Strong ante.\ & $10/14$  & $1/14$ & $1/14$ & $1/14$ & $1/14$ &  $0$ & 26.2 \\
Strong cons.\ & $1/14$  & $10/14$ & $1/14$ & $1/14$ & $1/14$ &  $42$ & 27.4 \\
\bottomrule		
\end{tabular}
\end{table}

In summary, we observed in this evaluation 
 that \algCossu{} is able to recover the planted rules under good conditions and that performance degrades as noise increases.

\subsection{Evaluation on a prediction task}
\label{ssec:exp_nextev}

In this experiment, our dataset consists of a life log posted on the site quantifiedawesome.com\footnote{\url{http://quantifiedawesome.com/records}} 
that records the daily activities of its owner, an enthusiast of the ``Quantified Self'' movement, since 2011. 
We downloaded the publicly available data for the period 11/2012--08/2019 (46445 events).
This dataset strikes an interesting balance between regularity in the activities (so that patterns can be learnt) 
and variations in the daily patterns (so that learning is far from trivial). Also, the interpretation of the rules
does not require special domain expertise. 
 
 \begin{figure*}[th]
    \begin{minipage}{.35\textwidth}
    \centering
    \includegraphics[width=\textwidth]{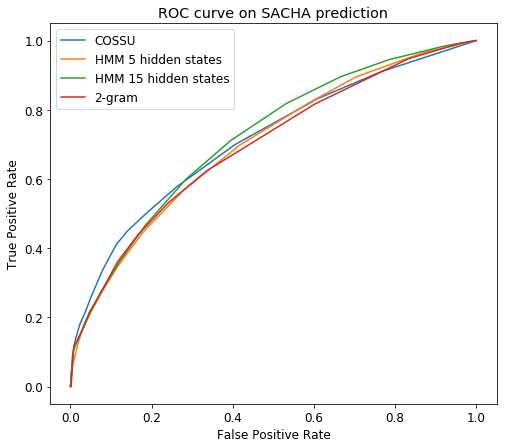}
    
    {\scriptsize
    \begin{tabular}{@{\hspace{1pt}}r@{\hspace{6pt}}r@{\hspace{1pt}}}
  \algCossu{}:~$0.7134$ & HMM \phantom{1}5:~$0.7040$ \\
  Bigram:~$0.6979$  & HMM 15:~$0.7166$ \\
    \end{tabular}}
    \end{minipage} \hfill
    \begin{minipage}{.64\textwidth}
    \centering
    \includegraphics[width=\textwidth]{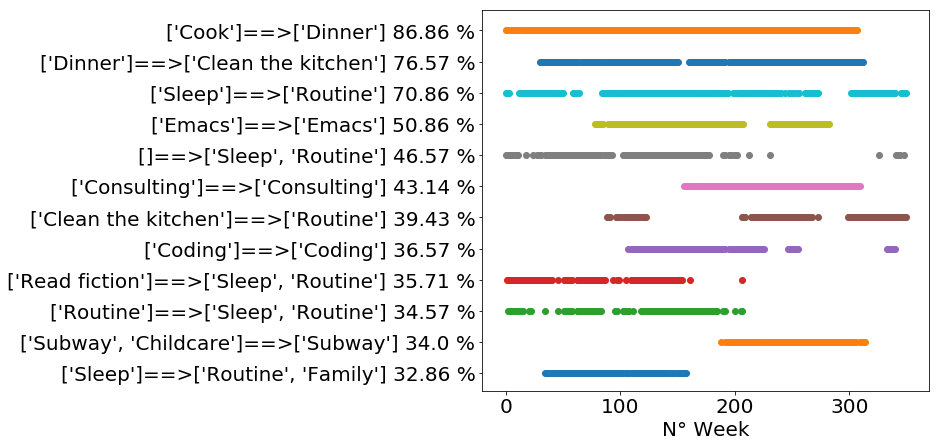}
    \end{minipage}
    \caption{ROC curves and corresponding Area Under Curve (AUC) comparing the prediction performance of \algCossu{} on the Quantified Awesome dataset, against two HMM with 5 and 15 hidden states, and a 2-gram predictor with different minimum confidence thresholds (left). Most frequent rules found by \algCossu{} on Quantified Awesome dataset (right).}
    \label{fig:rulesCossuSacha}
    \label{fig:rocSachaPrediction}
\label{tab:AUC}
\end{figure*}

\begin{table*}[th]
\caption{Results for next event prediction on the Quantified Awesome dataset. Note that the baseline prediction does not depend on a confidence threshold ($\tau$).}
\centering
\begin{footnotesize}
\begin{tabular}{@{\hspace{5pt}}p{0.08\textwidth}@{\hspace{5pt}}p{0.1\textwidth}@{\hspace{15pt}}p{0.07\textwidth}@{\hspace{2pt}}p{0.07\textwidth}@{\hspace{2pt}}p{0.07\textwidth}@{\hspace{2pt}}p{0.07\textwidth}@{\hspace{2pt}}p{0.07\textwidth}@{\hspace{5pt}}}
\toprule
  Method & $\tau$  &  \phantom{0}0 \% &  \phantom{0}20 \% &  \phantom{0}40 \% &  \phantom{0}60 \% &  \phantom{0}80 \% \\
\midrule
 \algCossu{} & Precision (\%) & $\phantom{0}26.67$ & $\phantom{0}38.57$ & $\phantom{0}56.93$ & $\phantom{0}73.69$ & $\phantom{0}85.05$ \\
  & Recall (\%) & $100.00$ & $\phantom{0}69.91$ & $\phantom{0}41.03$ & $\phantom{0}18.16$ & $\phantom{00}9.17$  \\
  & F1 score (\%) & $\phantom{0}42.11$ & $\phantom{0}49.71$ & $\phantom{0}47.69$ & $\phantom{0}29.14$ & $\phantom{0}16.56$  \\ [0.1em]
 HMM 5 & Precision (\%) & $\phantom{0}24.30$ & $\phantom{0}26.28$ & $\phantom{0}39.54$ & $\phantom{0}58.49$ & $\phantom{0}85.87$ \\
  & Recall (\%) & $100.00$ & $\phantom{0}96.51$ & $\phantom{0}56.73$ & $\phantom{0}21.13$ & $\phantom{00}4.83$ \\
  & F1 score (\%) & $\phantom{0}39.10$ & $\phantom{0}41.31$ & $\phantom{0}46.60$ & $\phantom{0}31.04$ & $\phantom{00}9.15$ \\[0.1em]

 HMM 15 & Precision (\%) & $\phantom{0}27.08$ & $\phantom{0}28.34$ & $\phantom{0}39.99$ & $\phantom{0}57.40$ & $\phantom{0}78.23$ \\
  & Recall (\%) & $100.00$ & $\phantom{0}98.45$ & $\phantom{0}71.18$ & $\phantom{0}27.60$ & $\phantom{0}11.98$ \\
  & F1 score (\%) & $\phantom{0}42.62$ & $\phantom{0}44.01$ & $\phantom{0}51.21$ & $\phantom{0}37.28$ & $\phantom{0}20.78$ \\[0.1em]

 Bigram & Precision (\%) & $\phantom{0}25.99$ & $\phantom{0}32.26$ & $\phantom{0}52.24$ & $\phantom{0}67.72$ & $\phantom{0}82.23$ \\
  & Recall (\%) & $100.00$ & $\phantom{0}81.68$  & $\phantom{0}35.95$ & $\phantom{0}15.69$ & $\phantom{0}11.74$ \\
  & F1 score (\%) & $\phantom{0}41.26$ & $\phantom{0}46.25$ & $\phantom{0}42.59$ & $\phantom{0}25.48$ & $\phantom{0}20.55$ \\ [0.3em]
  Baseline & \multicolumn{6}{l}{Precision (\%) $11.53$ \qquad Recall (\%) $100.00$ \qquad F1 score (\%) $20.68$} \\
 \bottomrule
\end{tabular}
\label{tab:nextEvtPredResults}
\end{footnotesize}
\end{table*}

\begin{table*}[th]
\centering
\caption{Basic statistics of our experimental datasets for classification (left) and accuracy of the different methods (right). The best performing method is in bold. \emph{Training sequence} is the length of the sequence used by \algCossu{} to learn the rules for the target classes $c_1$ and $c_2$. $|\Alphabet|$ is the number of different words after pre-processing the data. A dash (-) indicates that the classifier is not applicable to the data.}
\begin{tabular}{@{\hspace{5pt}}l@{\hspace{10pt}}r@{\hspace{20pt}}rr@{\hspace{10pt}}r@{\hspace{2.8em}}l@{\hspace{10pt}}l@{\hspace{10pt}}l@{\hspace{10pt}}l@{\hspace{5pt}}}
 \toprule
 Dataset & Nb.\ of  & \multicolumn{2}{c}{Training sequence} & $|\Alphabet|$  & \algCossu{} &  SVM  & HMM & BERT \\
  & instances & $c_1$ & $c_2$ & \\
 \midrule
 Presidential Debate  & 555 & 17658 & 20457 & 2414 & 0.798 & 0.779 & \textbf{0.803} & 0.731 \\ 
  Newsgroups & 180 & 13373 & 10895 & 5235 & 0.96 & 0.96  & 0.95 & \textbf{0.98} \\ 
  Critics & 1800 & 8722 & 8891 & 5558  &  \textbf{0.707}  & 0.697  & \textbf{0.707} & 0.705 \\ 
  Quant. Awesome & 1095 & 14562 & 5698 & 109 &  \textbf{0.72} & 0.59 & 0.69 & - \\  
  Stylometry & 2000 & 14273 & 17331 & 3966 & 0.72 & 0.7 & \textbf{0.75} &  \textbf{0.75}  \\  
  Stylometry (PoS) & 1200 & 11342 & 10629 & 31 & \textbf{0.64} & 0.59 & 0.605 & - \\  
 \bottomrule
\end{tabular}
\label{tab:realdt}
\label{tab:classification-results}
\end{table*}

\spara{Experimental setup.}
We divide the data into multiple train/test setups with one year (365 consecutive days) 
of training data and the immediately following week as testing data.
These train/test couples are built for each of the 350 such test weeks (starting on Mondays) for which the data is available. 
Each test week is a sequence of events and a prediction is made for every event of the sequence, 
considering as input the sequence of events up to the event to predict (excluded). 

To predict the next event in the sequence we used the rules mined by \algCossu{} to compute the probability of apparition of every symbol of the alphabet (as the inverse of the log of the symbol's code length). The symbol with the highest probability is predicted if its probability 
is higher than a given minimum confidence threshold $\tau$. 

\spara{Competitors.}
We compare \algCossu{} to methods that can both predict next events and provide some understanding about the data. 
These methods are: 
\begin{description}[topsep=3pt,leftmargin=0.5cm]
    \item[Baseline] predicts the same event as exactly one year before at this time; 
    \item[Bigram] predicts as follows: all pairs of events $aX$ (bigrams) are retrieved from the training data, and the frequencies of the different symbols $X$ following an $a$ are computed. The  highest frequency symbol is predicted if its frequency is higher than the minimum confidence threshold $\tau$.
    \item[Hidden Markov Models (HMM)] are trained on a one-year sequence and used to compute, for each symbol of the alphabet, the probability to be the next one. The highest probability symbol is predicted if its probability is above $\tau$. We test HMMs with 5 and 15 hidden states.
\end{description}

\spara{Results.}
We present the results at different confidence thresholds in Table \ref{tab:nextEvtPredResults}.

The F1 performance of the baseline is poor (20.68\%), and \algCossu{} is always better, except for $\tau=80\%$ where very few rules 
from \algCossu{} are allowed to make a prediction. This makes the recall plummet but gives one of the best precision values of the experiments. The table also shows that \algCossu{} outperforms bigrams, and is comparable to HMM with 15 hidden states. 
This is confirmed by the representation of the same results as a ROC curve shown in Figure \ref{fig:rocSachaPrediction} (which also provides the AUC of the curves). 
Overall, HMMs have higher recall, while \algCossu{} has higher precision.
This illustrates well the difference of the two approaches: an HMM learns a global model of the data that can cover a wider range of situations and is more robust to noise. 
On the downside, this global model is not interpretable. In contrast, \algCossu{}'s rules can be seen as a set of interpretable local models that are combined for prediction: they might do very well on some portions of the sequence but be less suitable on noisy portions. 

As an illustration, Figure \ref{fig:rulesCossuSacha} shows the top-10 rules most frequently selected by \algCossu{} during the training phases. 
For example, the rule $\textit{Cook} \rightarrow \textit{Dinner}$ has appeared in \algCossu{}'s output in 86.86\% of the training phases. 
A point is displayed for each rule on the test weeks where this rule was discovered during training (one year immediately before). 
We note that among rules identified by \algCossu{} some are applicable across the entire sequence and represent very mundane activities, such as $\textit{Cook} \rightarrow \textit{Dinner}$ or $\textit{Dinner} \rightarrow \textit{Clean\;the\;kitchen}$, whereas others are more specific to particular time periods. 
For example, in the year preceding week 200 in the graphic, the log's author gave birth to her first child. 
This drastically changed her daily habits, as can be seen for instance by the disappearance of the rule $\textit{Read\;fiction} \rightarrow \textit{Sleep, Routine}$, and the emergence of the rule $\textit{Subway, Childcare} \rightarrow \textit{Subway}$.

\subsection{Evaluation on a classification task}
In this experiment, we use the following datasets:
\begin{description}[topsep=3pt,leftmargin=0.5cm]
\item[Quantified Awesome] is the same dataset as in Section~\ref{ssec:exp_nextev}, for which we predict whether a sequence of events happened on a weekday or during the weekend;
\item[Presidential debates]\footnote{\url{https://www.kaggle.com/mrisdal/2016-us-presidential-debates}} consists of 555 sentences uttered by Donald Trump and Hillary Clinton in the context of the 2016 US election; the goal is to predict the speaker given a sentence;
    \item[Newsgroups]\footnote{\url{https://kdd.ics.uci.edu/databases/20newsgroups/20newsgroups.html}} is a dataset of 180 posts about the topics \emph{electronic} and \emph{religion}; 
    \item[Film critics]\footnote{\url{https://github.com/clairett/pytorch-sentiment-classification/tree/master/data/SST2}} is a dataset of 1800 movie reviews used for sentiment analysis;
    \item[Stylometry]\footnote{\url{https://bit.ly/3wBqITA}} is a database of 2000 writings from H.P. Lovecraft or E.A. Poe, where the goal is to predict the author based on its writing style; 
    \item[Stylometry (PoS)] is the sequence of part-of-speech tags from the previous dataset.
\end{description}
For all datasets (except Quantified Awesome), our alphabet $\Alphabet$ consists of all the words present in the corpus after removing special characters and stop words, and stemming of the remaining words. All datasets were split into training, validation, and test subsets. Table~\ref{tab:realdt} shows basic statistics of the datasets.

\spara{Experimental setup.} In line with the MDL-based classifier proposed in~\cite{krimp}, we can use \algCossu{} to classify sequences as follows: For each class $c$, we build a \emph{training sequence} by concatenating all the sequences labeled with $c$. We then extract sets of rules with \algCossu{} on the resulting long sequence. This set of rules defines a model $H_c$ that characterizes the generation process for sequences in class $c$. To predict the class of an input sequence $S$, we compute the description length $L(S|H_c)$ for each class $c$ and return the class that compresses $S$ best, i.e., the class that minimizes $L(S|H_c)$.

We compare the \algCossu{} classifier to (i) a SVM classifier trained on a bag-of-words representation of the text instances, (ii) an HMM-based classifier, and (iii) logistic regression trained on top of the BERT language model~\cite{bert}.

\spara{Results.} Classification accuracies on the test sets are reported in Table~\ref{tab:classification-results}. 
For all methods, the reported performance was obtained with hyperparameters tuned for best performance in cross-validation.
Despite not being a native classifier, \algCossu{} achieves the best performance on three of the experimental datasets, and exhibits comparable performance to HMM and state-of-the-art text classifiers. Furthermore, and unlike its competitors, \algCossu{} is inherently interpretable because its rules serve as an explicit explanation for the outcome of classification. As anecdotal examples, we show a few rules characterizing the classes of the Stylometry dataset:

\smallskip
 \noindent \begin{tabular}{@{\hspace{5pt}}l@{\hspace{.5em}}l@{\hspace{5pt}}}
  \textit{E.A.\ Poe:} & rue $\rightarrow$ morge \\ 
  & ourang $\rightarrow$ outang \\ 
  & little old $\rightarrow$ gentleman \\
  \textit{H.P.\ Lovecraft:} & pnakotic $\rightarrow$ manuscript \\ 
  & catch eight $\rightarrow$ clock coach arkham \\
  & necronomicon mad $\rightarrow$  arab abdul alhazred
 \end{tabular}

\section{Conclusion}
We have presented \algCossu{}, an algorithm to mine sequential rules from a long sequence of symbolic events. Our approach is inspired by the MDL principle and retrieves a compact and relevant set of rules. Experiments on real-world data show that, in addition to providing an interpretable model for sequential data, the retrieved rules achieve competitive accuracy on the tasks of next event prediction and a classification, as compared to other symbolic methods. 

As future work, we would like to design a procedure to optimize the weights of the rules while dynamically adjusting their precision.
We would also like to allow our method to handle gaps, perhaps by taking inspiration from \cite{sqs}. An implementation of the method is available at \url{https://gitlab.inria.fr/lgalarra/cossu}.

\spara{Acknowledgements.}
The authors would like to thank Nikolaj Tatti for providing a fast implementation of closed gapless sequential patterns miner.
    
\bibliographystyle{IEEEtran}
\let\clrp\clearpage
\let\clearpage\relax
\vspace{0.1cm}
\bibliography{references}
\let\clearpage\clrp

\end{document}